\documentclass{article}

\usepackage[accepted]{icml2020}

\usepackage{microtype}
\usepackage{graphicx}
\usepackage{subfigure}
\usepackage{booktabs} 

\usepackage{hyperref}

\usepackage[utf8]{inputenc} 
\usepackage[T1]{fontenc}    
\usepackage{url}            
\usepackage{booktabs}       
\usepackage{amsfonts}       
\usepackage{nicefrac}       
\usepackage{microtype}      
\usepackage{xspace}
\usepackage{amsmath}
\usepackage{accents}
\usepackage{multirow}
\usepackage[svgnames]{xcolor}

\newcommand{\nas}{{\sc NAS}\xspace}

\newcommand{\ws}{{\sc ws}\xspace}
\newcommand{\rs}{{\sc rs}\xspace}
\newcommand{\re}{{\sc re}\xspace}

\newcommand{\enas}{{\sc enas}\xspace}

\newcommand{\darts}{{\sc darts}\xspace}
\newcommand{\nao}{{\sc nao}\xspace}
\newcommand{\sota}{{\sc sota}\xspace}
\newcommand{\nasbench}{{\sc nasbench-\small{101}}\xspace}
\newcommand{\cifart}{{\sc cifar-\small{10}}\xspace}

\newcommand{\afour}{{\sc $\mathcal{A}_4$}\xspace}
\newcommand{\athree}{{\sc $\mathcal{A}_3$}\xspace}
\newcommand{\atwo}{{\sc $\mathcal{A}_2$}\xspace}
\newcommand{\aone}{{\sc $\mathcal{A}_1$}\xspace}
\newcommand{\afull}{{\sc $\mathcal{A}_{\text{full}}$}\xspace}
\newcommand{\afullres}{{\sc $\mathcal{A}_{\text{full}}^{\text{res}}$}\xspace}
\newcommand{\afullnores}{{\sc $\mathcal{A}_{\text{full}}^{\text{nres}}$}\xspace}

\definecolor{myred}{rgb}{0.8,0,0}
\definecolor{mygreen}{rgb}{0,0.7,0}
\definecolor{myblue}{rgb}{0,0,0.7}

\icmltitlerunning{To Share or Not To Share: An Extensive Appraisal of Weight-Sharing}

\begin{document}

\twocolumn[
\icmltitle{To Share or Not To Share: An Extensive Appraisal of Weight-Sharing}




\begin{icmlauthorlist}
\icmlauthor{Alo\"is Pourchot}{gleamer,isir}
\icmlauthor{Alexis Ducarouge}{gleamer}
\icmlauthor{Olivier Sigaud}{isir}
\end{icmlauthorlist}

\icmlaffiliation{gleamer}{Gleamer 96bis Boulevard Raspail, 75006 Paris, France}
\icmlaffiliation{isir}{Sorbonne Universit\'e, CNRS, Institut des Syst\`emes Intelligents et de Robotique, F-75005 Paris, France}

\icmlcorrespondingauthor{Alo\"is Pourchot}{alois.pourchot@gleamer.ai}

\icmlkeywords{Machine Learning, NAS, Neural Architecture Search, weight-sharing, ICML}
]

\vskip 0.3in



\printAffiliationsAndNotice{}  

\begin{abstract}
Weight-sharing (\ws) has recently emerged as a paradigm to accelerate the automated search for efficient neural architectures, a process dubbed Neural Architecture Search (\nas). Although very appealing, this  framework is not without drawbacks and several works have started to question its capabilities on small hand-crafted benchmarks. In this paper, we take advantage of the \nasbench dataset to challenge the efficiency of \ws on a representative search space. By comparing  a \sota \xspace \ws approach to a plain random search we show that, despite decent correlations between evaluations using weight-sharing and standalone ones, \ws is only rarely significantly helpful to \nas. In particular we highlight the impact of the search space itself on the benefits.
\end{abstract}

\section{Introduction}

Using deep neural networks (DNNs) has led to numerous breakthroughs on many hard  machine learning tasks, such as object detection and recognition or natural language processing \citep{lecun2015deep}. In the last years, a paradigm shift was observed, from hand-designing features that can be fed to a machine learning algorithm, to hand-designing neural architectures that can extract those features automatically. 
However, arranging DNNs is itself time-consuming, requires a lot of expertise and remains very domain-dependent. A promising approach is to automatically design them, a process referred to as Neural Architecture Search (\nas) \citep{nassurveyelsken2018,wistuba2019survey}.

Regrettably, because of expensive training requirements, evaluating a single DNN architecture can take days to weeks. In turn, original \nas approaches \citep{regularizedevoreal2018, transferablezoph2017, nasrlzoph2016} required thousands of GPU days worth of computing, only to find conformations slightly better than expert-designed ones. In light of this concern, many methods have been explored that could drastically cut the resources required to perform \nas, and today's literature is blooming with approaches requiring less than a day of computations \citep{enaspham2018, dartsliu2018, snasxie2018, casale2019pnas}

Most of these efficient methods rely on a computational trick called weight-sharing (\ws). First popularized by \citep{brock2017smash} and \citet{enaspham2018}, \ws proposes to reuse sets of weights from previously trained networks, rather than training each newly chosen architecture from scratch. \citeauthor{enaspham2018} push this idea further by noticing that, in their search space, each network can be seen as a sub-graph of a larger graph: the "super-net".
Using \ws therefore allows the training of the whole search space at once, by using a single set of weights (represented by the super-net), from which each possible model can then extract its parameters.

Despite a growing literature, the effects of \ws on the performances of \nas are still poorly understood. A particular concern is the quality of the scores obtained with the super-net. Employing \ws implies substituting metrics obtained after standalone training with metrics derived from the shared set of parameters. Both quantities thus need to be correlated: if networks with excellent standalone performances were under-evaluated with the super-net or vice-versa, the process could be pointless or even detrimental.

Studying this matter requires training many architectures and is itself extremely costly. As described in Section~\ref{sec:related}, several works mitigate this issue by assessing the correlations between evaluations of the super-net and true evaluations in a reduced setting, either evaluating few architectures or studying a drastically reduced search space. In this paper we leverage an existing dataset of architecture evaluations, \nasbench \citep{nasbench2019ying}, to investigate whether \ws can improve \nas in practice. 

Our experimental results show that: (i) with the correct methodology, one can get decent correlations between super-net proxy evaluations and real evaluations on several search spaces containing hundred thousands of architectures; (ii) despite correct correlations, \ws is inconsistent and rarely yields significant improvement over a random search (\rs) or robust search algorithms such as Regularized Evolution (\re) \cite{regularizedevoreal2018}; (iii) correlations might not be the limiting factor of \ws, as search spaces offering better correlations do not always offer better \ws performances; (iv) the search-space can bias training and evaluation of the super-net, which could explain the strong dependence of \ws performances over the search space.

\section{Related Work}
\label{sec:related}

In this section we describe several works trying to measure the efficiency of weight-sharing.

\citet{bender2018understanding} train a super-net on a search space of their own. Path dropout is applied during training to randomly zero-out some portions of the super-net. A simple random search is then used to find a good architecture. To validate the use of the super-net as a proxy to standalone accuracy, $20,000$ architectures are sampled from the chosen search space and evaluated with the resulting super-net. Then this set is partitioned into several bins based on the obtained proxy scores. For each bin, $4$ architectures are sampled and trained from scratch for a small number of epochs (around $10\%$ of the length of baseline training) before being evaluated. The authors note visually satisfying correlations between the two proxies, but do not report any numerical metrics. Correlations with full budget standalone accuracy are not reported, most likely for computational reasons. Moreover, because the few models evaluated are evenly spread across the range of possible proxy accuracies, the produced appealing correlations plots might not be representative of the whole search space.

\citet{evaluating2019sciuto} quantify the impact of \ws on a small language modeling task. They find that a simple random search baseline is competitive with and sometimes outperforms several \nas algorithms exploiting \ws such as \darts \citep{dartsliu2018}, \enas \citep{enaspham2018} and \nao \citep{luo2018nao}, which furthermore suffer from high variance. They report over a search space of $32$ architectures a poor correlation between the ranks obtained using a super-net and with standalone evaluations. Their study is however limited by the small size of the search space and the used algorithms, which were not specifically designed to produce good correlations at the end of training, but rather exploit them to rapidly converge to seemingly good architectures.

\citet{zhang2020deeper} explore another small search space of 64 architectures dedicated to computer vision. They train several super-nets using different seeds, and report high variance in the relative rankings of the architectures obtained with \ws. They notice that during super-net training, strong interactions exist between architectures, as updates in some models can either improve or deteriorate the performance of others. They reach correct correlations with standalone rankings, albeit the important variance seems to hinder the practical implications of the super-net. They propose several approaches to reduce the amount of \ws between architectures, such as fine-tuning parts of the super-net before evaluation or grouping architectures into different sets according to different strategies. 

\citet{chu2019fairnas} argue that \ws is limited by uneven sampling of individual weights throughout the learning process. Although they are seen equally often on average, some might locally be over-represented due to random chance, effectively biasing the weights of the super-net. To prevent this, they propose to average the gradient updates of the shared parameters over $n$ samples, chosen such that each of the $n$ elementary operations of the super-net appears exactly once in the resulting computational graphs. They combine their super-net trained with the aforementioned strategy with a multi-objective genetic algorithm, to search for a Pareto front of accurate architectures with adequate numbers of parameters and multiply-adds. They also report that, when sampling 13 models equally distant from the found Pareto front and comparing the accuracy of \ws vs standalone, the rankings are well preserved. However, details regarding this experiment are lacking and it is likely that the result do not hold for the whole search space.

\citet{luo2019improvingosnas} also note a strong variance in the results of a few \nas algorithms exploiting \ws. Using $50$ random models sampled from their search space, they evaluate the correlations between the scores given by the super-net and the scores obtained by training from scratch. They report poor correlations, which they deem responsible for the impaired results of \ws. After imputing the meager correlations to several factors such as short training times and bias towards simple architectures, they propose for each of them simple solutions that improve correlations.

Closest to us, \citet{zela2020nasbenchshot} use \nasbench to evaluate \nas algorithms exploiting \ws. However,  because the authors choose to study \darts variants, they create their own search spaces to perform evaluations, whereas we can directly use the whole \nasbench search space. In one of their experiments, they report the evolution during training of the correlations between evaluations obtained using the super-net, and evaluations queried from the \nasbench dataset. They report poor or nonexistent correlations for most algorithms, which seems to contradict our findings.

\section{Background}
\label{sec:background}

In this Section, we present the \nasbench dataset and the foundations of the \ws approach. We furthermore introduce some improvements to the standard super-net training that have been suggested in the literature.

\subsection{NASBench-101}

\label{sec:nasbench}
Assessing in practice the quality of any \nas approach is costly since it requires to evaluate all the architectures of a realistic search space. Fortunately, catalogues of such evaluations are starting to appear. In our experiments, we use the \nasbench dataset  \citep{nasbench2019ying} which matches $423,000$ unique architectures trained on \cifart to their training time, training accuracy, validation accuracy and test accuracy. This allows us to query the value of an architecture under constant time.

The \nasbench search-space is inspired by the one described in \cite{transferablezoph2017}, which is a standard one-shot \nas reference \citep{dartsliu2018, snasxie2018, casale2019pnas}. A global architecture consists of the successive iterations of a computational cell which optimal local architecture is to be found. Cells are represented by directed acyclic graphs where the first and last nodes correspond to the input and output. Other nodes represent applied operations. The flow of data itself is represented by the directed edges. At each active node, an operation is chosen among $3 \times 3$ max-pooling, $1 \times 1$ convolution and $3 \times 3$ convolution. We refer the reader to \cite{nasbench2019ying} for more details on the search space, their training and evaluation procedures, and how they split the \cifart dataset.

\subsection{Weight-Sharing}
\label{sec:ws}

Weight-sharing refers to the process of combining the weights of all the architectures of a search space into a single super-net. To access a model and its weights, one only needs to activate the corresponding computational sub-graph. The shared parameters are learned by successively activating different parts of the super-net and performing the standard forward-backward propagation algorithm on mini-batches of data. 
The optimization problem solved when performing \nas using \ws can be written as successive iterations of two steps:
\begin{align}
    \label{eq:ws}
    \text{Find} \quad &W \in \arg\min_{W \in \mathbb{R}^N} \Phi(\mathcal{A}, f, W), \\
    \text{Find} \quad &A \in \arg\min_{A \in \mathcal{A}} F(A, W),
\end{align}

where $\mathcal{A}$ is the set of possible architectures, $W$ the weights of the super-net, $F$ is the outer objective (usually a validation loss), $f$ is the inner objective (usually a training loss) and $\Phi$ is a function of the inner objective and the search space which dictates how to optimize $W$. The loss $\Phi$ is usually expressed as an expectation of the inner objective over a distribution $P_\theta$ of architecture:
\vskip -0.15in
\begin{equation}
    \label{eq:phi}
    \Phi(\mathcal{A}, f, W) = \mathbb{E}_{A \sim P_\theta(\mathcal{A})}\{f(A, W)\}.
\end{equation}

Different approaches combine both phases in different ways. Most alternate between the two, with mini-batches of data respectively coming from training and validation sets \cite{dartsliu2018, snasxie2018, casale2019pnas, enaspham2018}. In this work, we first train a super-net until convergence, and then use it to select possibly good architectures. We take as baseline the work of \cite{guo2019spuniform}, where models are sampled uniformly from the set of all possible architectures and the super-net is updated in accordance. This paradigm facilitates the analysis of the correlations, as methods that train both weights and architectures together induce a bias towards architectures with good early evaluations. Besides, \cite{guo2019spuniform} report better performance when exploiting their trained super-net to perform \nas.

\subsection{Enhancing Weight-Sharing Correlations}
\label{sec:improvements}

Several tricks and weight-sharing training variants have been introduced in the literature to improve the correlations offered by the super-net. We list several unrelated approaches here, which we explore in our experiments in Section~\ref{sec:ranking} and \ref{sec:ranking_results}.

During evaluation, it is possible to directly exploit the whole super-net and perform a standard forward pass on the impending data whilst activating the graph corresponding to the evaluated net. However, several works report the benefit of adapting the statistics of the inherited batch normalization layers \citep{bender2018understanding, guo2019spuniform}.

The weights of the super-net $W$ are updated through gradient descent with respect to the objective in \eqref{eq:ws}, using the formulation of $\Phi$ described in \eqref{eq:phi}. The resulting gradient takes the form of an expectation over the distribution $P_\theta$: 
\begin{align}
\label{eq:trux}
    \nabla_W \Phi(A, f, W) = \mathbb{E}_{A\sim P_\theta(\mathcal{A})}\{\nabla_W f(A,W)\}.
\end{align}
This expectation is approximated by an empirical average, using random architectures sampled following $P_\theta(\mathcal{A})$. However, in practice \cite{guo2019spuniform} only use a single architecture to estimate the expectation. Although this process is unbiased, it results in high variance updates of $W$. Decreasing this variance by sampling more models could improve the super-net optimization, at a higher computational cost.

In \cite{stamoulis2019spnas}, the authors propose to not only share the weights of the basic operations between architectures, but to further merge the weights of all basic operations at a given node into a single set of parameters. For instance, if two basic operations were a $x\times x$ convolution and a $y\times y$ convolution were $x>y$, then instead of representing both operations with two different sets of kernels, one could use a single set of kernels of size $x\times x$, and apply the $y\times y$ convolution by extracting the sub-kernels of size $y\times y$ from the bigger ones.

\cite{luo2019improvingosnas} identify in their work a bias towards architectures with fewer free parameters, as they are easier to train than more complex ones. They propose to correct this bias by sampling architectures pro-rata to their number of parameters, resulting in more complex architectures being sampled more often.  

\section{Experimental Study}
\label{sec:experimental_study}
In this section, we describe the protocols used to address the following questions: Do accuracies of architectures obtained with \ws correlate with standalone ones? Do we get the same correlations under various training and evaluation regimes? Can \ws consistently outperform random search? Do the results vary between search spaces? The outcomes are described in Section~\ref{sec:results}\footnote{The complete codebase is available at \url{https://github.com/apourchot/to_share_or_not_to_share}}.

\subsection{Ranking Architectures with Weight-Sharing}
\label{sec:ranking}

We want to establish the achievable correlations between the accuracies obtained with a super-net, and the accuracies obtained after standalone trainings.
We train several super-nets on the \cifart dataset. Following \citet{guo2019spuniform}, for each mini-batch of data seen during training, a single architecture is uniformly sampled from the search space. The weights of the super-net are then updated according to the computational graph generated by the activation of this architecture. We reuse the hyper-parameters of \citet{nasbench2019ying}, which we detail in the supplementary document. The only notable differences are that we reduce the initial number of filters from 128 to 16, and train networks for four times longer to let the super-nets converge.

Our choice to reduce the number of initial filters to 16 is motivated by computational reasons. According to earlier iterations of our experiments, using the default 128 value requires fairly longer training times for the super-nets, without significantly improving the resulting correlations. Although fixing it to 16 is somewhat arbitrary, \cite{zela2020nasbenchshot} also note in their work that accuracies obtained after training architectures with 16 initial filters are greatly correlated with those obtained using the baseline 128 filters. We thus consider that the number of filters is not the limiting factor of our different \ws experiments. Furthermore, this setup mimicks one-shot \nas approaches such as, \cite{dartsliu2018, casale2019pnas, enaspham2018}, where the model found by the search is often up-scaled to further improve accuracies.

We train 5 different super-nets on each search space. Then we randomly sample 1,000 unique\footnote{See the supplementary document for a note on this matter.} architectures from the search space and compute their proxy accuracies on a held-out validation dataset. We match those accuracies with the average validation accuracies returned by \nasbench. To quantify the quality of the correlations between the two, we use Spearman's rank correlation coefficient.

We estimate accuracies with the super-nets performing either no fine-tuning ({\sc no-ft}) or fine-tuning the batch-norm statistics ({\sc bns-ft}). Without fine-tuning, we directly use all the parameters and batch-norm statistics of the super-net. When adapting the batch-norm statistics of the super-net to a specific architecture, we average them over 4 mini-batches of data.

Additionally, we study the effect on correlations of the different training variations mentioned in Section~\ref{sec:improvements}. We follow the same protocol and always fine-tune the batch-norm statistics ({\sc bns-ft}). We consider four variants: averaging the gradients in Equation~\eqref{eq:phi} over 3 architectures ({\sc avg-$3$)}, sampling architecture pro-rata to their number of parameters ({\sc pro-rata}) \citep{luo2019improvingosnas}, following the single-kernel approach of \citeauthor{stamoulis2019spnas} ({\sc single-k}), and combining the single-kernel and pro-rata approaches ({\sc s-k + p-r}). 

\subsection{Impact of Weight-Sharing on NAS Performances}
\label{sec:nas}
Quantifying the correlations obtained with \ws on realistically sized search-spaces is interesting as such, but it is not enough to conclude on the efficiency of \ws itself. Indeed, it is not clear above what correlation level \ws becomes useful. Here we aim at characterizing the interest of substituting super-net evaluations to the standalone evaluations when performing \nas. To investigate this, we perform \nas using three different approaches: a control random search (\rs), the Regularized Evolution (\re) algorithm \citep{regularizedevoreal2018} and random search guided by a super-net. We report the evolution of the corresponding test regrets as a function of time. This test regret is computed after each model evaluation by comparing the mean test accuracy of the model with the running best validation accuracy and the best mean test accuracy of the considered search space. 

For \rs, we evaluate $10,000$ unique architectures. For the \re algorithm, we reuse the implementation and the hyper-parameters provided by \cite{nasbench2019ying} and stop search after evaluating $10,000$ valid models (models that don't belong to the search space are directly given an accuracy of 0). When exploiting \ws, we also consider $10,000$ randomly sampled unique architectures. However, instead of evaluating them in a random order, we assess their performances using a trained super-net and query \nasbench in decreasing order of proxy accuracy, under the batch-norm statistics fine-tuning ({\sc bns-ft}) setting. The \re baseline is already known to perform better than RS in the long-term on \nasbench \citep{nasbench2019ying}, and helps quantify the regret differences between \rs and \ws.

The different regrets are plotted as a function of search time. This search time is well approximated by the duration of the training and evaluation of the different models, as both phases account for most of the duration of a search.
For a fair time-wise comparison of the regrets, we accounted in the \ws-based method for both super-nets training time and the duration of the evaluation of the pool of models. Unfortunately, we were not able to perform experiments using the same hardware as \citeauthor{nasbench2019ying}. As a result, using training times reported in \nasbench would be biased, as theirs was much more efficient. To circumvent this, we used a common setup comprised of two NVidia K80 GPUs to estimate for each distinct architecture the time required to perform a single forward and a single backward pass when using 128, and 16 initial feature maps. We averaged this quantity over three independent measures. To get the length of training a model, we then simply multiply the resulting quantity by the appropriate number of times both passes are performed. Besides, given that during  super-net training each sampled architecture only activates the necessary parts of the network, we approximated the super-net training time  by the average training time of the architectures of the considered search spaces. 

Our comparison is focused on the one-shot \nas paradigm, where the super-net is used to choose a few good models which are then re-trained from scratch. We consider selecting $1$, $10$ and $20$ architectures, and respectively refer to those strategies as {\sc top-$1$}, {\sc top-$10$}, and {\sc top-$20$}. Given a search space, we estimate the average regret $\bar{\mathcal{R}}_{ws}$ reached by exploiting each {\sc top-x} strategy. We then compute the average regret achieved by \rs and \re \emph{under the same time budget}, respectively $\bar{\mathcal{R}}_{rs}$, and $\bar{\mathcal{R}}_{re}$, and report the absolute differences in average regret between \rs and the \ws-guided strategy, and between \rs and \re. To get an insight on the effect size, we also report the Cohen's $d$, which is defined as the average difference divided by the pooled standard deviation. Cohen's $d$  \cite{cohen1988statistical} reflects how the measured mean absolute difference relates to the standard deviations of both populations. An effect is usually considered important if $|d| > 0.8$, mild if $|d|\approx0.5$ and small if $|d|<0.2$. 

We assess the statistical significance of the absolute differences using a Student's $t$-test. For \rs, and \re, since there is almost no computational requirements (because a run is a simple query of the \nasbench dataset), we follow \cite{nasbench2019ying} and perform $500$ runs. However, running a \ws-based search takes a non-negligible amount of time. Given our computational budget, we settle on ensuring that any effect of at least medium size can be properly measured, with a statistical power of $\beta=0.8$ and a significance level $\alpha=0.05$. We estimated using the {\sc statsmodels} python package \citep{seabold2010statsmodels} that $30$ runs of the \ws-based approach should grant such guarantees.



\subsection{Impact of Search Spaces on NAS Performances}
\label{sec:sp}

\begin{figure}[!t]
  \centering
    \includegraphics[width=0.97\linewidth]{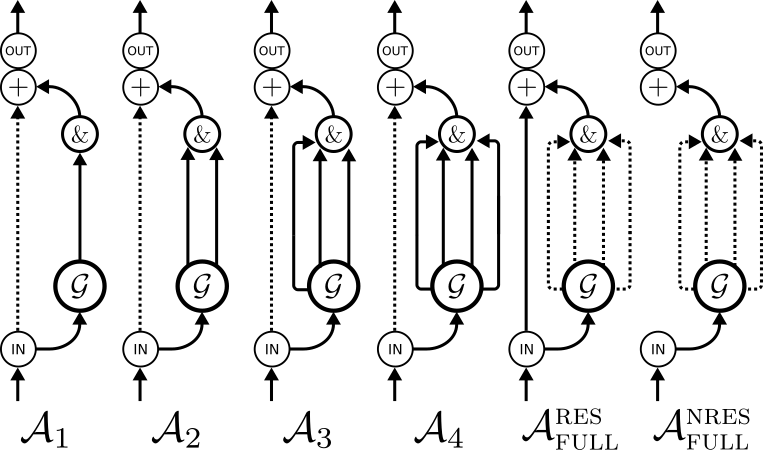}
    
   \caption{Structural properties of the different search-spaces. On each graph, "{\sc in}" and "{\sc out}" denote the input and output of the cell, while "$+$" and "$\&$" denote the sum and the concatenation of incoming features maps. We only display the edges discriminating at least one search space and represent the rest of the graph with node $\mathcal{G}$. If an edge does not discriminate a specific search space, we represent it with a dotted line. \aone to \afour are characterized by the number of edges concatenated after $\mathcal{G}$, and \afullres and \afullnores by the presence or absence of a residual connection. \label{fig:sps}}
   \vspace{-0.6cm}
\end{figure}

\begin{table*}[!th]
\caption{\label{table:corrs} Spearman's rank correlation coefficient between \ws and standalone evaluations for various search-spaces, \ws variants, and evaluation schemes. We report the average over $5$ independent runs and the 95\% confidence interval for its estimation. On the left, we use the baseline \ws approach and the three evaluation schemes described in Section~\ref{sec:ws}. On the right, we test some variants of \ws described in Section~\ref{sec:improvements} and always fine-tune batch-norm statistics during evaluations. Results marked with an asterisk {\small $*$} indicate that one of the super-net failed to converge, and that the reported statistics are computed using only the four others. 
}
\begin{center}
\begin{small}
\begin{sc}
{\renewcommand{\arraystretch}{1.2}
\begin{tabular}{l | r | r || r | r | r | r  }
\toprule
 & 
 \multicolumn{1}{c}{\phantom{{\small $*$}} no-ft} & 
 \multicolumn{1}{c||}{\phantom{{\small $*$}} bns-ft} & 
 \multicolumn{1}{c}{\phantom{{\small $*$}} single-k} & 
 \multicolumn{1}{c}{\phantom{{\small $*$}} pro-rata} &
 \multicolumn{1}{c}{\phantom{{\small $*$}} avg-$3$} &
 \multicolumn{1}{c}{\phantom{{\small $*$}} s-k + p-r}
 \\
\midrule
\afour  & 0.08 $\pm$ 0.17 & \textbf{0.64} $\pmb{\pm}$ \textbf{0.03} & 0.66 $\pm$ 0.01 & 0.68 $\pm$ 0.03 & {\small $*$} 0.67 $\pm$ 0.02& \textbf{0.69} $\pmb{\pm}$ \textbf{0.02} \\
\athree  & 0.12 $\pm$ 0.15 & \textbf{0.59} $\pmb{\pm}$ \textbf{0.03} & 0.62 $\pm$ 0.02 & 0.63 $\pm$ 0.02& 0.61 $\pm$ 0.03& \textbf{0.66} $\pmb{\pm}$ \textbf{0.02} \\
\atwo & 0.24 $\pm$ 0.03 & \textbf{0.60} $\pmb{\pm}$ \textbf{0.04} & 0.64 $\pm$ 0.02 & 0.64 $\pm$ 0.02& 0.61 $\pm$ 0.01& \textbf{0.65} $\pmb{\pm}$ \textbf{0.02} \\
\aone & 0.32 $\pm$ 0.05 & \textbf{0.68} $\pmb{\pm}$ \textbf{0.02} & 0.72 $\pm$ 0.02 & \textbf{0.75} $\pmb{\pm}$ \textbf{0.02}& 0.73 $\pm$ 0.01& 0.67 $\pm$ 0.01 \\
\afull & 0.24 $\pm$ 0.05 & \textbf{0.56} $\pmb{\pm}$ \textbf{0.04} & {\small $*$} \textbf{0.63} $\pmb{\pm}$ \textbf{0.02}& \phantom{{\small $*$}} 0.59 $\pm$ 0.03 & \phantom{{\small $*$}} 0.61 $\pm$ 0.02 & \phantom{{\small $*$}} 0.58 $\pm$ 0.02\\
\afullnores & {\small $*$} 0.11 $\pm$ 0.05 & {\small $*$} \textbf{0.46} $\pmb{\pm}$ \textbf{0.06} & {\small $*$} \textbf{0.58} $\pmb{\pm}$ \textbf{0.02} & \phantom{{\small $*$}} 0.56 $\pm$ 0.03 & \phantom{{\small $*$}} 0.52 $\pm$ 0.02 & \phantom{{\small $*$}} 0.49 $\pm$ 0.02\\
\afullres & 0.34 $\pm$ 0.10 & \textbf{0.71} $\pmb{\pm}$ \textbf{0.02} & 0.68 $\pm$ 0.01& \textbf{0.72} $\pmb{\pm}$ \textbf{0.02}& 0.69 $\pm$ 0.02& 0.66 $\pm$ 0.01\\
\bottomrule
\end{tabular}}
\end{sc}
\end{small}
\end{center}
\end{table*}

To measure the impact of the search space on \ws, we introduce several sub-sets of the \nasbench search space. 

In \nasbench, feature maps going to the output of a cell are concatenated. However, since the size of the output is fixed across all possible cells,
the input size, output size, and number of filters of nodes may vary over architectures, possibly hindering the use of \ws. A reasonable splitting strategy is to consider the sets $(\mathcal{A}_i)_{i=1,\dots,4}$, in which architectures contain exactly $i$ nodes connected to the output (beside the input node, which is always added), and therefore share as many parameters and feature maps\footnote{A difference of $1$ feature map can still appear in \athree since the number of final feature maps after concatenation is rarely divisible by $3$. In such case, one branch may end up with one more or one less feature map, e.g. $[42, 43, 43]$ for $128$ output feature maps}.
We also consider the full \nasbench set, which we call \afull: we solve the aforementioned problem by dynamically adapting the number of feature maps used for each node depending on the architecture: each node, as seen by the super-net, contains the maximum number of filters for the given layer, but sampled architectures only inherit the first $n$ filters of the filter-bank, where $n$ is determined so as to satisfy the constraints on the output size the architecture's cell. 

Early results additionally compelled us to study the influence of residual connections on \ws. It is well known in the computer vision literature that edges connecting the input node to the output node, known as residual connections \citep{he2015resnet}, significantly influence the quality of the optimization of individual architectures. Suspecting that this is also true when training super-nets, we consider two additional search spaces: \afullres and \afullnores, respectively containing all architectures of \nasbench with and without residual connections. Figure~\ref{fig:sps} sketches the structural properties of the different search spaces.

\section{Results}
\label{sec:results}
We now describe the results of the above studies. We then discuss the influence of the search space on \ws.

\subsection{Ranking Capabilities of Weight-Sharing}
\label{sec:ranking_results}

For each search space, we report in the left part of Table~\ref{table:corrs} the average over $5$ super-nets of the rank correlation between the average standalone accuracies returned by \nasbench and the proxy accuracies obtained after applying the two evaluation protocols described in Section~\ref{sec:ranking}. {\sc no-ft} refers to performing no fine-tuning, and {\sc bns-ft} to fine-tuning the batch-norm statistics. Without any fine-tuning, we get the worst correlations across all search-spaces, with substantial variance. With batch-norm statistics fine-tuning, the average correlation increases by $270\%$ over the {\sc no-ft} scheme, giving 3 times betters results on average. 


In the right part of Table~\ref{table:corrs}, we present the rank correlations obtained from training with the different variants of \ws described in Section~\ref{sec:improvements} and \emph{fine-tuning batch-norm statistics during evaluations}. {\sc single-k} refers to applying the single kernel variant \citep{stamoulis2019spnas}, {\sc pro-rata} to sampling architectures pro-rata to their number of parameters \citep{luo2019improvingosnas}, and {\sc s-k + p-r} the combination of the two. {\sc avg-{\footnotesize 3}} refers to averaging gradients over three architectures. We notice that all approaches lead to a small improvement of the correlations, as well as a slight variance reduction.

These simple results show that, \emph{as long as batch-norm statistics are adapted to the evaluated architectures, it is possible to get correlations between proxy evaluations performed with \ws, and full-budget evaluations}. We notice  that all the works mentioning poor correlations in Section~\ref{sec:related} do not detail their evaluation setup, and we suspect that they do not adapt batch-norm statistics. Additionally, \emph{it is possible to further improve the resulting correlations by slightly modifying the super-net training in different ways.}

\subsection{Can Weight-Sharing Improve NAS ?}
\label{sec:nas_results}

The regret curves of the different NAS experiments described in Section~\ref{sec:nas} are reported in \figurename~\ref{fig:regrets}. As can be observed from the results, the \ws-guided strategy is almost always close to either \rs, or \re and only once significantly below the two, on \aone.

We report in Table~\ref{table:pvals} with color codes the average regret difference between \rs and \re and between \rs and the \ws-guided policy for the {\sc top-1} and {\sc top-10}  paradigms. For space reasons, results from the {\sc top-20} paradigm are reported in the supplementary document. Numerical and statistical results coincide with the visual results of \figurename~\ref{fig:regrets}. 

Using the {\sc top-$1$} scenario, applying \ws results in a $\emph{smaller}$ ($d< -0.5$, red) regret on \aone and \afull, a $\emph{slightly smaller}$ ($-0.5 <d< -0.2$, light red) regret on \atwo, \afullres and \afullnores, a slightly $\emph{larger}$ regret ($0.2<d<0.5$, light blue) on \athree, and a larger regret ($d>0.5$, blue) on \afour. On the other hand, \re produces $\emph{smaller}$ regrets on all searches but \afull, where it is roughly $\emph{equivalent}$ to \rs, and on \afullres, where it returns $\emph{larger}$ regrets on average. 

Using the {\sc top-$10$} paradigm, applying \ws results in a smaller regret on \aone, \afull, \afullres and \afullnores and a slightly smaller regret on \atwo. On \afour and \athree the $p$-values suggest no statistical significance of the slightly $\emph{larger}$ regrets for the \ws-based approach, which hints at a performance roughly equivalent to that of \rs, as can be observed from \figurename~\ref{fig:regrets}. \re produces smaller regrets across all searches but \aone on which it is only slightly better, and \afullres, where it is again roughly equivalent to \rs. 

\begin{table*}[!th]

\xdefinecolor{lblue}{named}{DodgerBlue}
\xdefinecolor{blue}{named}{Blue}
\xdefinecolor{lred}{named}{Salmon}
\xdefinecolor{red}{named}{Red}

\caption{\label{table:pvals} For the {\sc top-1} and {\sc top-10} paradigms, we report the average regret difference between \rs and the \ws-guided strategy, and between \rs and \re as well as the 95\% confidence intervals for the estimation of the mean difference. \textbf{For clarity purposes, reported regrets are multiplied by 100}. We test for the statistical significance of the difference using an independent $t$-test and report the resulting $p$-values. We also report the Cohen's $d$ as a measure of the effect size. Results high-lighted in blue correspond to larger regrets ($d>0.5$), in light-blue to slightly larger regrets ($0.2<d<0.5$), in light-red to slightly smaller regrets ($-0.5<d<-0.2$) and in red to smaller regrets ($d<-0.5$). Results left blank indicate small effect sizes ($|d|<0.2$) or non-significant results ($p\geq0.05$). We additionally report in bold two outlying results.}
\begin{center}
\begin{scriptsize}
\begin{sc}
{\renewcommand{\arraystretch}{1.3}
\begin{tabular}{l|r|r|r|r}
\toprule
 & \multicolumn{2}{c|}{{\sc top-1}} & \multicolumn{2}{c}{{\sc top-10}} \\ \cmidrule[0.5pt]{2-5}
 & \multicolumn{1}{c|}{$\bar{\mathcal{R}}_{rs}$ $-$ $\bar{\mathcal{R}}_{re}$} & \multicolumn{1}{c|}{$\bar{\mathcal{R}}_{rs}$ $-$ $\bar{\mathcal{R}}_{ws}$} & \multicolumn{1}{c|}{$\bar{\mathcal{R}}_{rs}$ $-$ $\bar{\mathcal{R}}_{re}$} & \multicolumn{1}{c}{$\bar{\mathcal{R}}_{rs}$ $-$ $\bar{\mathcal{R}}_{ws}$} \\ \midrule
                                
\afour & 
\textcolor{red}{-0.47 $\pm$ 0.08 ($p$<0.01, $d$=-0.72)} & 
\textcolor{blue}{0.68 $\pm$ 0.37 ($p$<0.01, $d$=\phantom{-}0.90)} &
\textcolor{red}{-0.27 $\pm$ 0.06 ($p$<0.01, $d$=-0.56)} & 
0.07 $\pm$ 0.19 ($p$=0.50, $d$=\phantom{-}0.13) \\

\athree & 
\textcolor{red}{-0.39 $\pm$ 0.09 ($p$<0.01, $d$=-0.54)} & 
\textcolor{lblue}{0.33 $\pm$ 0.28 ($p$=0.03, $d$=\phantom{-}0.43)} & 
\textcolor{red}{-0.21 $\pm$ 0.05 ($p$<0.01, $d$=-0.51)} & 
0.15 $\pm$ 0.23 ($p$=0.23, $d$=\phantom{-}0.31)\\
                                 
\atwo & 
\textcolor{lred}{-0.38 $\pm$ 0.11 ($p$<0.01, $d$=-0.41)}& 
\textcolor{lred}{-0.41 $\pm$ 0.27 ($p$<0.01, $d$=-0.47)}&
\textcolor{red}{-0.21 $\pm$ 0.05 ($p$<0.01, $d$=-0.57)}&
\textcolor{lred}{-0.17 $\pm$ 0.09 ($p$<0.01, $d$=-0.41)}\\
                                 
\aone & 
\textcolor{lred}{-0.43 $\pm$ 0.18 ($p$<0.01, $d$=-0.29)}&
\textcolor{red}{\textbf{-1.37 $\pmb{\pm}$ 0.14 ($\pmb{p}$<0.01, $\pmb{d}$=-1.16)}}&
\textcolor{lred}{-0.10 $\pm$ 0.04 ($p$<0.01, $d$=-0.32)}&
\textcolor{red}{-0.24 $\pm$ 0.10 ($p$<0.01, $d$=-0.62)}\\

\afull &
0.11 $\pm$ 0.19 ($p$=0.24, $d$=\phantom{-}0.07)&
\textcolor{red}{-0.88 $\pm$ 0.17 ($p$<0.01, $d$=-0.85)}&
\textcolor{red}{-0.23 $\pm$ 0.05 ($p$<0.01, $d$=-0.58)}&
\textcolor{red}{-0.24 $\pm$ 0.12 ($p$<0.01, $d$=-0.52)}
\\
                                               
\afullnores&
\textcolor{red}{-0.63 $\pm$ 0.16 ($p$<0.01, $d$=-0.50)}&
\textcolor{lred}{-0.54 $\pm$ 0.42 ($p$=0.02, $d$=-0.47)}&
\textcolor{red}{-0.44 $\pm$ 0.06 ($p$<0.01, $d$=-0.98)}&
\textcolor{red}{-0.55 $\pm$ 0.12 ($p$<0.01, $d$=-0.96)}
\\ 
                                                               
\afullres&
\textcolor{blue}{\textbf{1.01 $\pmb{\pm}$ 0.22 ($\pmb{p}$<0.01, $\pmb{d}$=\phantom{-}0.56)}}&
\textcolor{lred}{-0.22 $\pm$ 0.13 ($p$<0.01, $d$=-0.39)}&
0.01 $\pm$ 0.04 ($p$=0.42, $d$=\phantom{-}0.05)&
\textcolor{red}{-0.17 $\pm$ 0.09 ($p$<0.01, $d$=-0.56)}\\

\bottomrule
\end{tabular}}
\end{sc}
\end{scriptsize}
\end{center}
\end{table*}

Results for the {\sc top-20} paradigm are similar to those of the {\sc top-10} paradigm, although all effects seem to have reduced amplitude.

On average across all search spaces the performances of \re constitute a strong baseline, even under severe time constraints, and \re looks more reliable than the \ws -based approach, except on \afullres, where it seems to outperform \rs only very late in the search, as \figurename~\ref{fig:regrets} shows. The results suggest nonetheless that \ws can improve the performance of random search but that its efficiency is inconsistent and on average relatively poor. To re-contextualize the different reported difference in regrets and the Cohen's $d$ measures, one can consider that with an effect size $d=-0.39$ for \ws over \rs (on \afullres under the {\sc top-1} paradigm), the probability that a random run with \ws produces a smaller regret than random search constrained with the same time budget is only 61\%\footnote{An interactive visualization of the phenomenon can be studied at \url{https://rpsychologist.com/d3/cohend/} \cite{magnusson2020}}. On \aone, where \ws is somehow very effective and produces a large effect size of $d=-1.16$, this probability reaches a maximum of 89\%. For a mild effect size of $d=-0.52$ (on \afull under the {\sc top-10} paradigm), it is around 65\%. As it has been noted several times in the literature \cite{li2019random, yang2019nasfrustrating, evaluating2019sciuto}, reporting the results over several runs is thus crucial to \nas research, especially when effect sizes are small. 

Moreover, Table~\ref{table:pvals} suggests no clear link between the level of correlation reached by \ws on a search space and its ability to outperform RS: On \afull, where correlations in Table~\ref{table:corrs} are the lowest, \ws significantly outperforms RS under the {\sc top-$10$} paradigm, whereas it offers terrible results on \afour despite significantly better correlations. \ws offers similar correlations on $\mathcal{A}_2$ and $\mathcal{A}_3$, but respectively smaller and larger regrets than RS. On \afullres and \aone, where \ws offers the best correlations, the \ws-guided search is respectively slightly better and much better than \rs.

Under the time constraints of one-shot \nas, \emph{\ws can slightly outperform \rs, but rarely to a great extent, and can even be worse}. Given the same time budget, the well-established \re algorithm is a more consistent improvement over \rs. Besides, there seems to be \emph{no obvious relationship between the level of correlation between proxy and standalone evaluations, and the performances of \ws on a search space.}

\subsection{Variations between Search Spaces}
\label{sec:sp_res}

\begin{figure*}[!thb]
 \centering
    \includegraphics[width=0.24\linewidth]{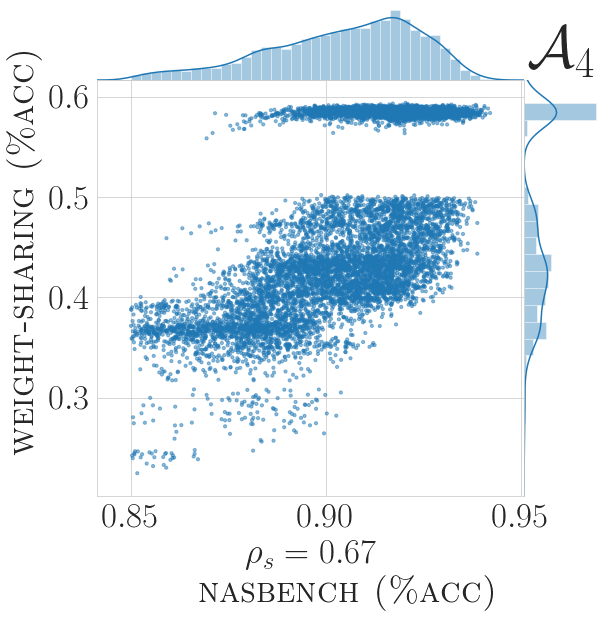}
    \includegraphics[width=0.24\linewidth]{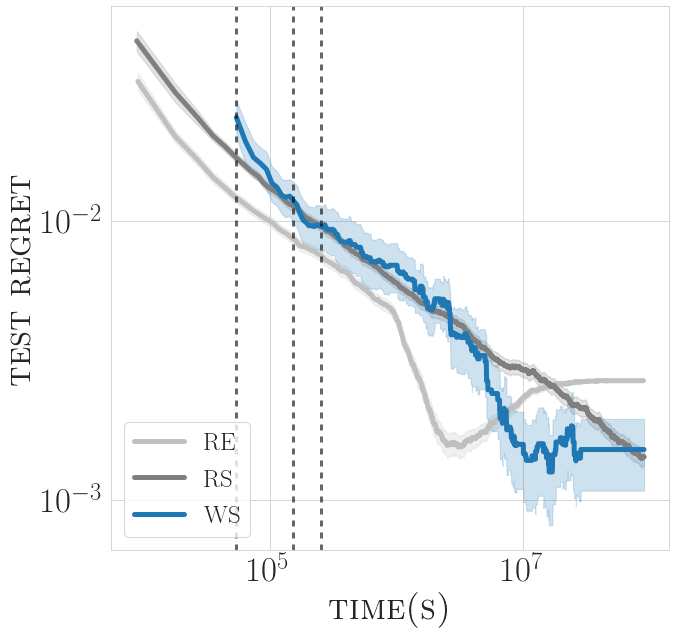}
    \includegraphics[width=0.24\linewidth]{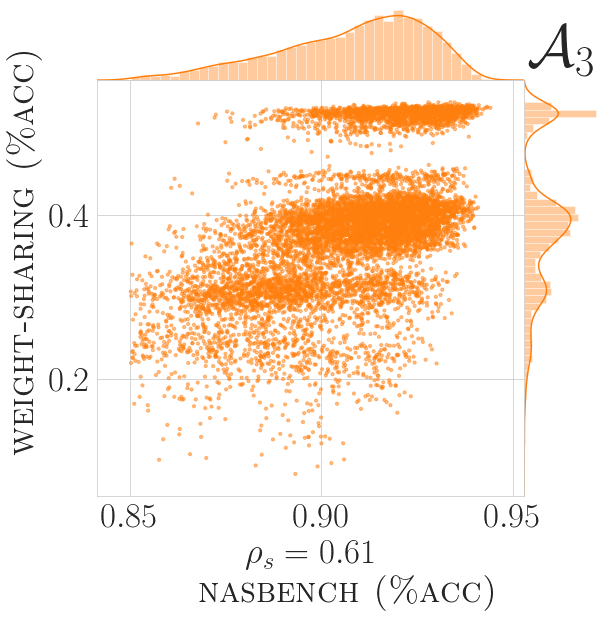}
    \includegraphics[width=0.24\linewidth]{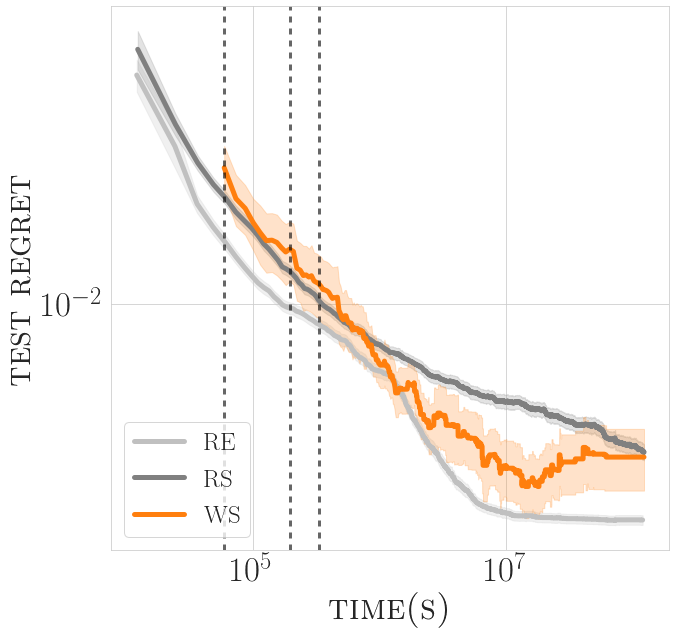}
    \includegraphics[width=0.24\linewidth]{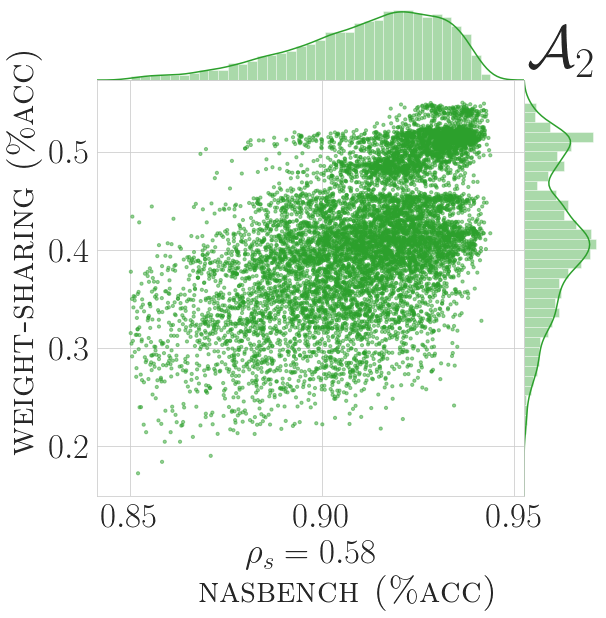}
    \includegraphics[width=0.24\linewidth]{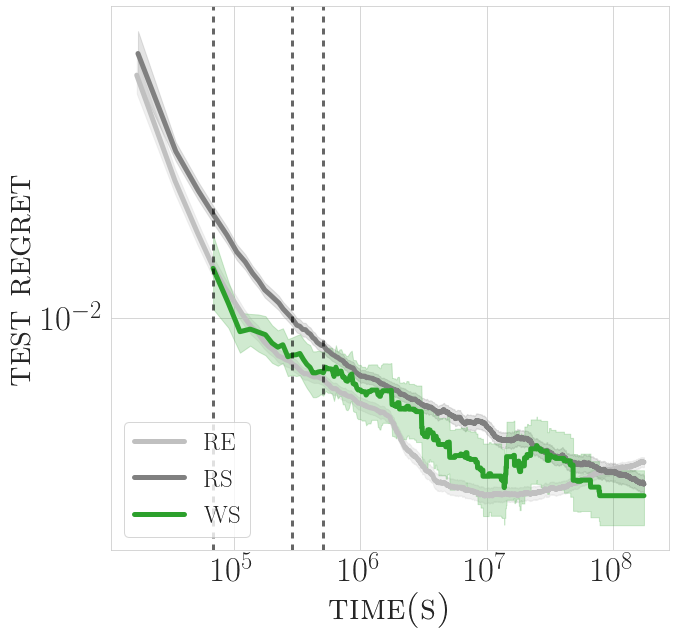}
    \includegraphics[width=0.24\linewidth]{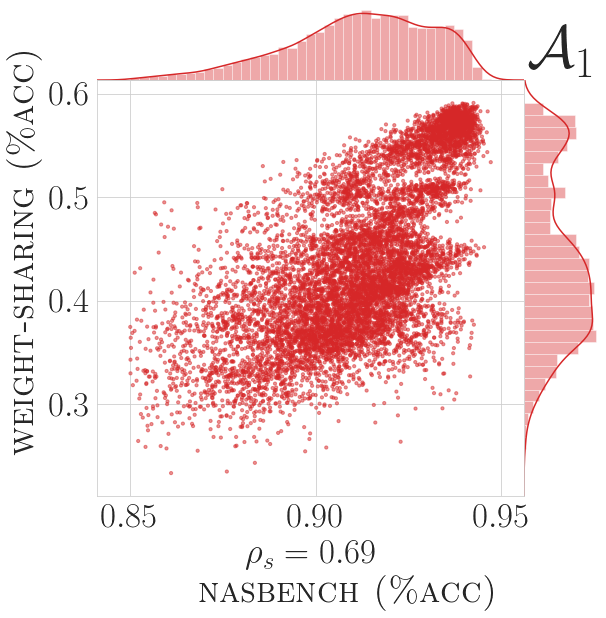}
    \includegraphics[width=0.24\linewidth]{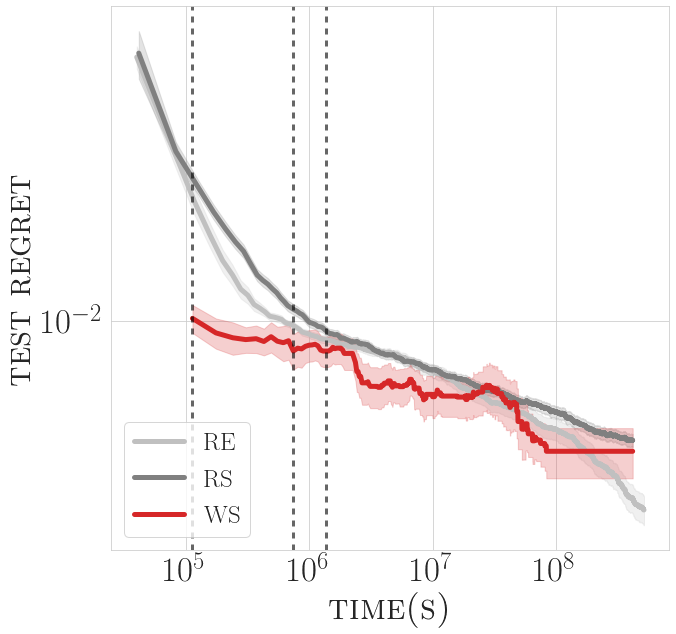}
    \includegraphics[width=0.24\linewidth]{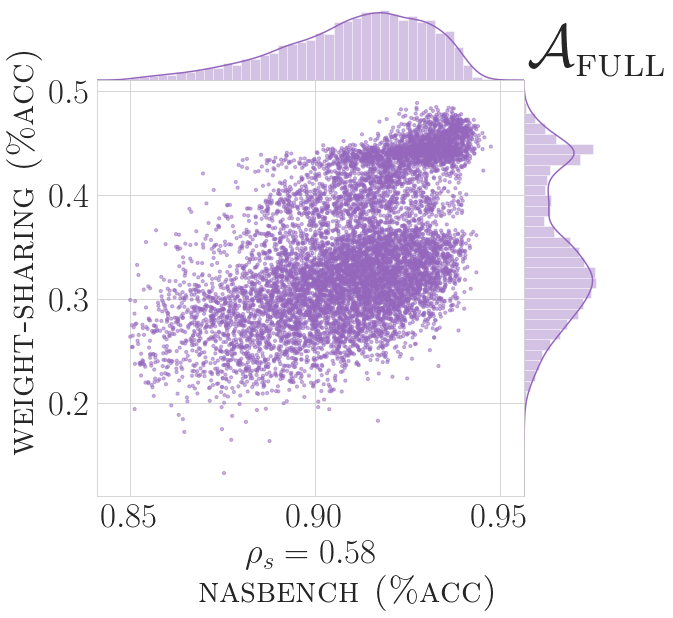}
    \includegraphics[width=0.24\linewidth]{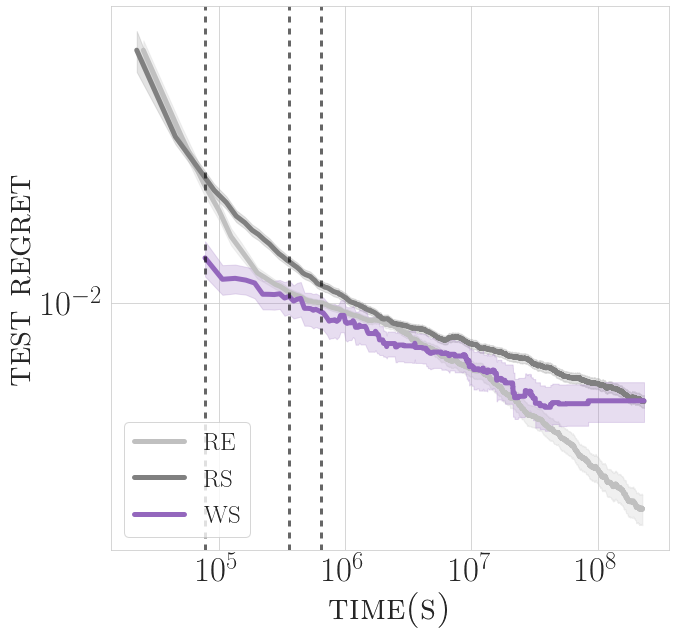}
    \includegraphics[width=0.24\linewidth]{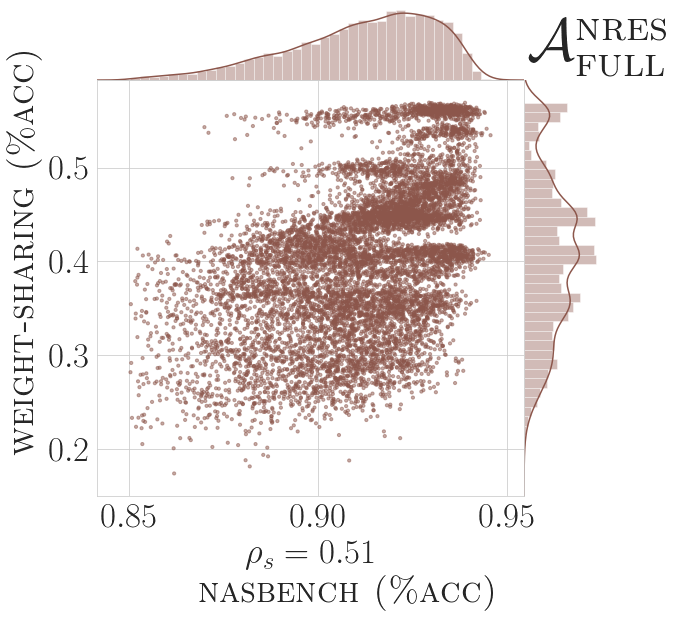}
    \includegraphics[width=0.24\linewidth]{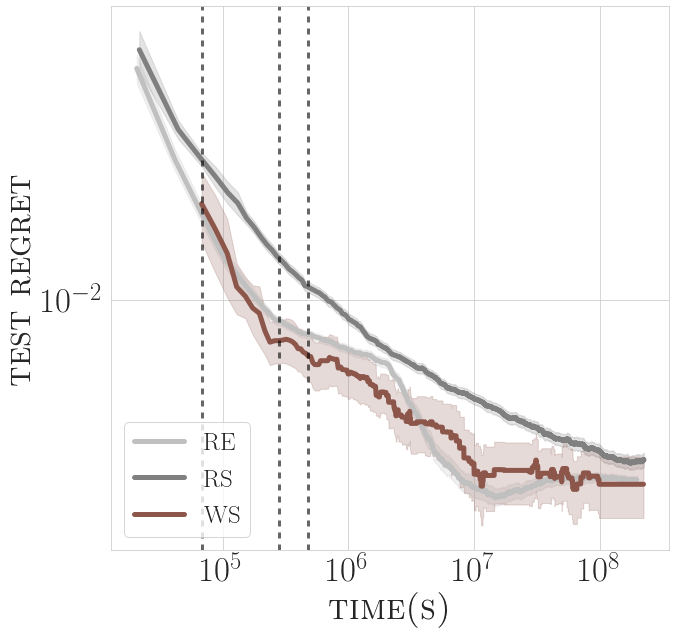}
    \includegraphics[width=0.24\linewidth]{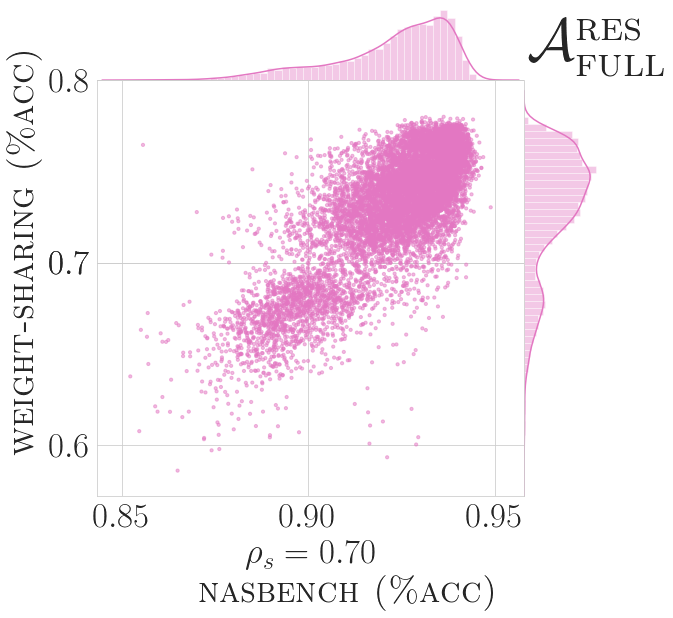}
    \includegraphics[width=0.24\linewidth]{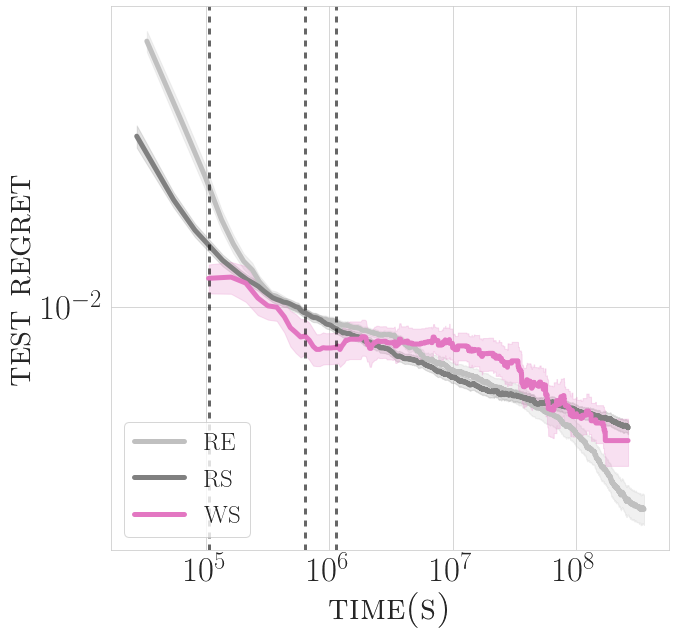}
    
   \caption{For each search space, we report on the left a scatter plot of the proxy accuracy computed using a super-net ({\sc bns-ft}) (y-axis), and the average validation accuracy returned by \nasbench (x-axis) for $10,000$ architectures. On the right, we report the evolution of the test regret as a function of time for the different \nas algorithms considered. Colored curves correspond to the \ws-based strategy, light-gray ones to \re and dark-gray ones to \rs. Curves are averaged over 500 runs for \rs and \re, and 30 runs for \ws. Visible colored areas correspond to the 95\% confidence interval for the estimation of the average. Notice that both axes use a logarithmic scale. Vertical lines correspond, from left to right, to the time-budget associated with respectively the {\sc top-1}, {\sc top-10} and {\sc top-20} paradigms.
   \label{fig:regrets}}
\end{figure*} 

From Section~\ref{sec:nas_results}, \ws-guided \nas seems to often slightly outperform RS, but this depends on the search space.

Coincidentally, we notice from the results of Section~\ref{sec:ranking_results} that simply changing the number of nodes connected to the output makes the average correlation vary between $0.59$ on $\mathcal{A}_3$ and $0.68$ on $\mathcal{A}_1$. Additionally, restricting the search space to architectures presenting a residual connection has a noticeable positive effect on the correlations, as they increase from $0.46$ to $0.71$ between \afullnores and \afullres. \emph{The search space itself has an important impact on the correlations, even more so than the training enhancements described in Section~\ref{sec:improvements}.}

The size of the datasets could explain the varying correlations. It has often been asserted in the literature that , the more architectures there are in the search space, the harder it is to train the super-net. The Spearman rank's correlation between the average correlation obtained with batch-size fine-tuning ({\sc bns-ft}) reported in Table~\ref{table:corrs} and the sizes of the dataset reaches $-0.71$ ($p=0.07$). The effect hints that larger search-spaces could possibly lead to smaller correlations between proxy and standalone evaluations, but the relatively low number of search spaces of this study prevents us from positively rejecting the null hypothesis that it does not with great confidence, and further studies are required to conclude on this matter. Besides, results in Section~\ref{sec:ranking_results} suggest that it is probably not the only aspect of the search space that is of influence. On $\mathcal{A}_2$ and $\mathcal{A}_3$, \ws offers roughly the same level of correlation, despite $\mathcal{A}_2$ being twice larger than $\mathcal{A}_3$. The correlation achieved is $25\%$ smaller in $\mathcal{A}_0^\emptyset$ than in $\mathcal{A}_0$, with $23\%$ less architectures. It is also interesting to note that few architectures are actually seen during training: given $432$ training epochs of $157$ mini-batches of data, less than $67,824$ unique architectures are used to update the super-net. This might be enough to cover \afour or \afullres, but represents only a tiny fraction of larger datasets, such as \atwo ($\simeq200,000$ architectures), or \afull ($\simeq400,000$ architectures). \emph{Further studies are required to clearly establish whether the size of the dataset has a non-negligible impact on the correlation capabilities of \ws, but several facts suggest that it cannot entirely explain the discrepancies between the different search spaces.}

We display in the supplementary document the five scatter plots obtained for each search space between the true validation accuracies and the proxy accuracies resulting from  from 5 different super-nets. We also report one for each search space in \figurename~\ref{fig:regrets}. There is a noticeable variance in the visual appearance of the figures, which is corroborated with the variance in the correlation coefficients reported in Table~\ref{table:corrs}. The numerical variance has been observed in several other studies of the literature \citep{evaluating2019sciuto, luo2019improvingosnas, zela2020nasbenchshot, zhang2020deeper} and is often attributed to the reliance of super-net training on sampled architectures. Interestingly, several visible clusters seem to be linked to proxy evaluations. For each scatter-plot, we report the distributions of the true validation and proxy accuracies over sampled architectures. Coincidentally with the different visible architecture clusters, distributions of proxy evaluations are much less regular than their true validation counterparts, often presenting several modes. \emph{The clusters of architectures in the scatter-plots visually transcribe existing biases in proxy evaluations}. 

\begin{figure}[!t]
\vskip 0.1in
  \centering
    \includegraphics[width=0.49\linewidth]{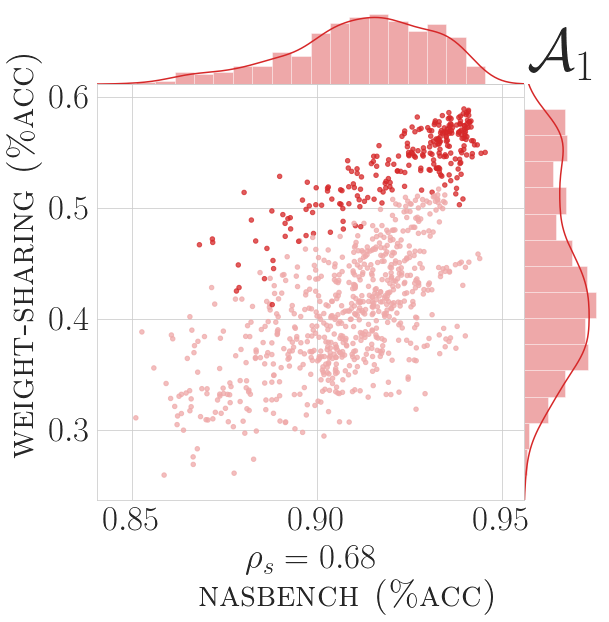}
    \includegraphics[width=0.49\linewidth]{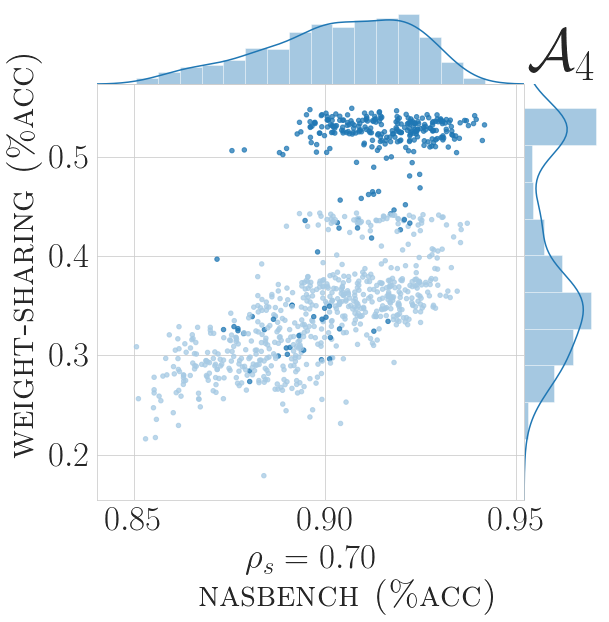}
    
   \caption{We report for a super-net trained on \aone (left) and \afour (right) the proxy accuracy computed after fine-tuning the batch-norm statistics (y-axis), and the average validation accuracy returned by \nasbench (x-axis) for $1,000$ architectures. We highlight in a darker tone the points corresponding to architectures with residual connections (left) and architectures with a $3\times3$ convolution on the first node (right). Both examples reveal a clear bias in super-net evaluations. We also report the distributions of the proxy and standalone accuracies of the sampled architectures.  \label{fig:bias}}
\end{figure}

There is no trivial relation between different biases and particular structural properties of the architectures. Fortunately, some biases are easier to highlight than others. We focus on two such biases in \figurename~\ref{fig:bias}. On \aone, architectures with a residual connection tend to get better evaluations than those without. On \afour, the presence of a $3\times3$ convolution on the first node triggers over-evaluation. Such clusters can be seen in the scatter plots of all search spaces except \afullres. \emph{Different search spaces bias the super-nets in different ways, resulting in different structural patterns of over/under-evaluations.}

The patterns appearing in the scatter-plots may explain the search results of Section~\ref{sec:nas_results} better than the correlations level reached by \ws. On $\mathcal{A}_4$, the over-evaluation bias visible in \figurename~\ref{fig:bias} creates a cluster of architectures with excellent proxy accuracies. As a result, \ws neglects a large number of architectures with equal or better capabilities that random search does not miss. Although the cluster contains a few of the best architectures, its average standalone accuracy is particularly poor. This impedes \ws from selecting good top-models, and makes the early \ws-guided search worse than random search. On $\mathcal{A}_1$, the over-evaluation bias towards residual connections benefits to the search, as architectures with residual connections are better on average and constitute most of the best architectures of the search space. The \ws-guided search is in turn quite efficient. \emph{The patterns of over/under-evaluations dictate the search behavior when exploiting \ws. If \ws is biased towards interesting patterns in the considered search space, then it is likely to outperform random search. Otherwise, the difference may not be significant. In the worst scenario, the bias can even be strong enough to undermine the performance of \ws.}

\section{Conclusion}

In this paper we have leveraged the \nasbench dataset to investigate the impact of weight-sharing on neural architecture search. Our results lead to the following conclusions.
First, super-nets trained with \ws can offer significant correlations between proxy evaluations and standalone evaluations, but fine-tuning the batch-norm statistics of the models is mandatory for the process to be successful. The results can be further improved by tweaking the \ws training process, but the search space itself has a more significant influence over the quality of correlations. 

More importantly, \ws is not consistently faster than a random search baseline, the improvement being mostly search-space dependent, and less reliable than those of well-established methods such as \re. Super-nets resulting from optimization with \ws can be biased towards specific structural patterns in the architectures, which also vary depending on the search-space. Those patterns, rather than the level of correlations, seem to dictate the efficiency of NAS when exploiting \ws. Given that each search space has its own specific biases, it is hard to foresee how well \ws is going to perform. Understanding in what ways the search space can bias the training of the super-nets emerges as a central question for the \ws paradigm and as a promising lead for future work. 

One limitation of this study is that we focus on the \nasbench dataset. It is unclear how the described results would transfer from \cifart to larger image datasets such as {\sc imagenet}. Further study will be required to investigate this matter, possibly with the help of new \nas datasets, such as {\sc nasbench-201} \cite{dong2020nasbench201}.

\bibliography{bib}
\bibliographystyle{icml2020}


\end{document}